\documentclass[review]{elsarticle}

\usepackage{lineno,hyperref}
\usepackage{diagbox}
\usepackage{amsfonts,amsmath}
\usepackage{multirow}
\usepackage[small]{caption}
\usepackage{graphicx}
\usepackage{slashbox}
\biboptions{sort&compress,comma}
\modulolinenumbers[5]

\usepackage{algorithm}  
\usepackage{algorithmicx} 
\usepackage{algpseudocode}  
\usepackage{caption}

\journal{Journal of \LaTeX\ Templates}

\begin{document}

\begin{frontmatter}

\title{Few-Shot Learning-Based Human Activity Recognition}


\author[mymainaddress]{Siwei Feng}
\ead{siwei@umass.edu}

\author[mymainaddress]{Marco F. Duarte\corref{mycorrespondingauthor}}
\cortext[mycorrespondingauthor]{Corresponding author}
\ead{mduarte@ecs.umass.edu}

\address[mymainaddress]{Department of Electrical and Computer Engineering, \\University of Massachusetts Amherst, \\Amherst, MA, 01003}

\begin{abstract}
Few-shot learning is a technique to learn a model with a very small amount of labeled training data by transferring knowledge from relevant tasks. In this paper, we propose a few-shot learning method for wearable sensor based human activity recognition, a technique that seeks high-level human activity knowledge from low-level sensor inputs. Due to the high costs to obtain human generated activity data and the ubiquitous similarities between activity modes, it can be more efficient to borrow information from existing activity recognition models than to collect more data to train a new model from scratch when only a few data are available for model training. The proposed few-shot human activity recognition method leverages a deep learning model for feature extraction and classification while knowledge transfer is performed in the manner of model parameter transfer. In order to alleviate negative transfer, we propose a metric to measure cross-domain class-wise relevance so that knowledge of higher relevance is assigned larger weights during knowledge transfer. Promising results in extensive experiments show the advantages of the proposed approach.
\end{abstract}

\begin{keyword}
Human Activity Recognition, Few-Shot Learning, Knowledge Transfer, Cross-Domain Class-Wise Relevance, Deep Learning
\end{keyword}

\end{frontmatter}


\section{Introduction}
\label{intro}

Studies in human activity recognition (HAR) have been attracting increasing attentions in recent years due to their potential in applications such as human health care \cite{lin2015agent}, indoor localization \cite{xu2016indoor}, smart hospital \cite{sanchez2008activity}, and smart home \cite{wen2016adaptive}. HAR is a technique that aims at predicting participants' activities from the low-level sensor inputs. In recent years, the use of wearable devices such as smart wristbands and smart phones has significantly facilitated HAR research due to their small sizes and low power consumption \cite{hu2018novel} as well as their capability of real-time information capturing \cite{saeedi2018signal}. \par
Traditional machine learning (ML) methods such as support vector machines and decision trees have significantly promoted the development of HAR studies during the past few decades. However, limitations such as heavy reliance on human domain knowledge \cite{bengio2013deep} and shallow feature extraction impede these ML methods from providing satisfactory results in most daily HAR tasks \cite{wang2018deep}. More recently, deep learning (DL) methods have been continuously providing marvelous performance in many areas such as computer vision and natural language processing \cite{lecun2015deep}. The capability to automatically extract high-level features makes DL methods largely alleviated from the drawbacks of conventional ML methods. There have been many DL-based HAR works proposed in recent years \cite{zeng2014convolutional, almaslukh2017effective, radu2016towards, hammerla2016deep}. However, the training time and the amount of data required for DL systems are always much larger than that of traditional ML systems, and the large time and labor costs makes it difficult to build a large-scale labeled human activity dataset with high quality. Therefore, it is necessary to find a way to alleviate these problems of DL-based HAR. \par
Few-shot learning (FSL) \cite{li2003bayesian, lim2011transfer, koch2015siamese, vinyals2016matching, snell2017prototypical, qi2018low} is a type of transfer learning technique \cite{pan2010survey} aiming at learning a classifier to recognize unseen classes (\textbf{target} domain\footnote{The definition of domain is introduced in Section \ref{rela}}) with only a small amount of labeled training samples by reusing knowledge from existing models on relevant classes (\textbf{source} domain), which makes it easier to train deep learning models with only a few training samples from each class. FSL has promising potential in many HAR applications. For example, after a health care system is installed, the system should be able to be retrained as the training data used in the previous system may not be representative of the new environment (e.g. new activity types, different people, etc.). Furthermore, the retraining should only require a small amount of new data, as users are generally not able to provide large amounts of data for training, especially for elderly or disabled people. Nonetheless, it may be helpful to keep some of the previously existing knowledge if we can discern between knowledge that is \emph{relevant} or helpful to learning the new tasks from the knowledge that is \emph{not relevant}, or that \emph{may even harm the learning performance} if used during training, a phenomenon known as negative transfer. \par
In this paper, we propose a novel FSL-based HAR method that we dub few-shot human activity recognition (FSHAR). The framework of FSHAR includes three steps. We first train a deep learning model with source domain samples, where the model parameters are randomly initialized. After that, we calculate the cross-domain class-wise relevance based on embeddings of both source domain samples and target training samples obtained from the source feature extractor. By leveraging the parameters from the source feature extractor and the classifier, as well as the obtained cross-domain class-wise relevance, we initialize the parameters for a target model. Fine-tuning is then performed on the initialized target model for final optimization. To the best of our knowledge, this is the first application of deep learning-based few-shot learning to human activity recognition.  \par
The key contributions of this paper are as follows:
\begin{itemize}
\item We propose a novel parameter transfer-based few-shot learning scheme for human activity recognition.
\item We propose a general framework to measure cross-domain class-wise relevance for human activity recognition to alleviate negative transfer.
\item We design a deep learning framework to transform the low-level sensor input of human activity signals to the high-level semantic information for human activity recognition.
\item We provide multiple experimental results to demonstrate the performance improvements achieved by the proposed method compared with relevant techniques from the state of the art.
\end{itemize}
The rest of this paper is organized as follows. Section \ref{rela} introduces notations used in this paper and overviews relevant techniques. Details of the proposed framework are presented in Section \ref{work}. Experimental results and the corresponding analysis are provided in Section \ref{exp}. Section \ref{con} concludes this work with suggestions for future work.

\section{Related Work}
\label{rela}

In this section, we introduce the paper’s notations and definitions and provide a review of techniques relevant to our proposed method. \par
Vectors are denoted by bold lowercase letters while matrices are denoted by bold uppercase letters. The superscript $T$ of a matrix denotes the transposition operation. For a matrix $\mathbf{A}$, $\mathbf{A}^{(q)}$ denotes the $q^{\mathrm{th}}$ column and $\mathbf{A}_{(p)}$ denotes the $p^{\mathrm{th}}$ row, while $\mathbf{A}^{(p,q)}$ denotes the entry at the $p^{\mathrm{th}}$ row and $q^{\mathrm{th}}$ column. The $\ell_{r,p}$-norm for a matrix $\mathbf{W} \in \mathbb{R}^{a \times b}$ is denoted as
\begin{equation}
\label{lrpnorm}
\|\mathbf{W}\|_{r,p} = \left( \sum_{i=1}^a \left( \sum_{j=1}^b |\mathbf{W}^{(i,j)}|^r \right)^{p/r} \right)^{1/p}.
\end{equation}
\par
We use $\mathbf{X} = [\mathbf{X}_{(1)}; \mathbf{X}_{(2)}; \cdots ; \mathbf{X}_{(n)}] \in \mathbb{R}^{n \times d}$ to denote sample sets, where $\mathbf{X}_{(i)} \in \mathbb{R}^d$ is the $i^{\mathrm{th}}$ sample in $\mathbf{X}$ for $i = 1, 2, \cdots , n$, and where $d$ and $n$ denote data dimensionality and number of samples in $\mathbf{X}$, respectively. \par
For notations of transfer learning, we use $\mathcal{D}$ to denote a domain and $\mathcal{T}$ for a task. A domain $\mathcal{D}$ consists of a feature space $\mathcal{X}$ and a marginal probability distribution $P(\mathbf{X})$ over a sample set $\mathbf{X}$. A task $\mathcal{T}$ consists of a label space $\mathcal{Y}$ and an objective predictive function $f(\mathbf{X}, \mathbf{Y})$ to predict the corresponding labels $\mathbf{Y}$ of a sample set $\mathbf{X}$. We use $\mathcal{D}_{\rm src} = \{ \mathcal{X}_{\rm src}, P(\mathbf{X}_{\rm src}) \}$ and $\mathcal{T}_{\rm src} = \{ \mathcal{Y}_{\rm src}, f(\mathbf{X}_{\rm src}, \mathbf{Y}_{\rm src}) \}$ to denote the source domain and task, and use $\mathcal{D}_{\rm trg} = \{ \mathcal{X}_{\rm trg}, P(\mathbf{X}_{\rm trg}) \}$ and $\mathcal{T}_{\rm trg}= \{ \mathcal{Y}_{\rm trg}, f(\mathbf{X}_{\rm trg}, \mathbf{Y}_{\rm trg}) \}$ for the target domain and task.

\subsection{Human Activity Recognition}

Human activity recognition (HAR) is a technique that aims at learning high-level knowledge about human activities from the low-level sensor inputs \cite{wang2018deep}.  In recent years, the growing ubiquity of sensor-equipped wearables such as smart wristbands and smartphones have significantly promoted researches regarding HAR in the field of pervasive computing \cite{bulling2014tutorial}. \par
Machine learning (ML) algorithms have been widely used in HAR. In the last decade, traditional ML tools such as Markov models \cite{zhu2009human, lee2011semi} and decision trees \cite{jatoba2008context, ermes2008advancing} have yielded tremendous progress in HAR. However, most traditional ML algorithms applied in HAR use manually designed features, which are shallow and heavily rely on human domain knowledge and experience; furthermore, they are specific to particular tasks. Therefore, traditional ML algorithms cannot handle complex HAR scenarios, and they require one specifically designed model for each task, which increases the time and labor cost to build HAR systems in terms of both labeled data collection and model construction. \par
In recent years, the application of deep learning (DL) methods to HAR has significantly alleviated the drawbacks of traditional ML based HAR methods. First, DL models can extract high-level features with little or no human design. Second, DL models can be reused for similar tasks, which makes HAR model construction more efficient. Different DL models such as deep neural networks \cite{walse2016pca, vepakomma2015wristocracy}, convolutional neural networks \cite{zeng2014convolutional, ha2015multi}, autoencoders \cite{almaslukh2017effective, wang2016human}, restricted Boltzmann machines \cite{radu2016towards, zhang2015real}, and recurrent neural networks \cite{inoue2018deep, edel2016binarized} have been applied in HAR. We refer readers to \cite{wang2018deep} for more details on DL-based HAR.

\subsection{Few-Shot Learning}
Few-shot learning (FSL) is a transfer learning technique that applies knowledge from existing data to data from unseen classes which do not have sufficient labeled training data for model training. In this paper we focus on the scenario of $\mathcal{X}_{\rm src} = \mathcal{X}_{\rm trg}$ while $\mathcal{Y}_{\rm src} \cap \mathcal{Y}_{\rm trg} = \emptyset$. \par
The first work for FSL is \cite{li2003bayesian}, in which a variational Bayesian framework is proposed to represent visual object categories as probabilistic models where existing object categories are used as the prior knowledge, while the model for unseen categories is obtained by updating the prior with one or more observations. Lim et al. \cite{lim2011transfer} propose a sample-borrowing method for multiclass object detection that adds selected samples from similar categories to the training set in order to increase the number of training data. \par
In recent years, deep learning based FSL \cite{koch2015siamese, vinyals2016matching, snell2017prototypical, qi2018low} has become the mainstream of FSL due to their unparalleled performance. Koch et al. \cite{koch2015siamese} proposed a double-network structure based on the deep convolutional siamese network to extract features from image pairs and generates a similarity score between inputs. Vinyals et al. \cite{vinyals2016matching} proposed matching networks to map a small labelled support set from unseen classes and an unlabelled example to its label. Snell et al. \cite{snell2017prototypical} proposed prototypical networks that learn a metric space to perform classification by computing distances to prototype representations of each class. Qi et al. \cite{qi2018low} proposed a weight imprinting schedule to add a weight for each new class into a softmax classifier. \par
During our literature search, we found that FSL is widely used in computer vision studies, motivated by the fact that human beings are able to recognize previously unseen objects with only a few training samples. By contrast, the application of FSL in HAR has been much more limited, especially when combined with DL. During our literature review, we did not find any DL based FSL method that is used in HAR.

\subsection{Long Short-Term Memory Network}

A long-short term memory (LSTM) network \cite{hochreiter1997long} is a type of recurrent neural network (RNN) which processes time series signals by taking as their input not just the current inputs but also what they have processed earlier in time. Each RNN contains a loop (repeating modules) inside the network structure that allows information to be passed from one step of the network to the next. \par
LSTMs are famous for their capabilities to capture long-term dependencies. Compared with the simple repeating modules of most RNNs, which sometimes only contains a single tanh layer, the repeating modules of LSTMs include more complicated interacting layers in their structures. The workflow of LSTM can be briefly described as follows. The first step is to determine the importance of previous information, which is to decide a status between ``completely forget about this" and ``completely keep this". The next step is to decide what new information to store in the cell state and then replace the old state with a new state. Finally, we decide the information to output. \par
A stacked LSTM model \cite{dyer2015transition} is a LSTM model with multiple hidden LSTM layers where each layer contains multiple memory cells. By stacking LSTM hidden layers, a LSTM model can be deeper that makes it capable of tackling more complex problems.

\section{Proposed Method}
\label{work}

\subsection{Basic Framework}
\label{framework}

\begin{figure}[t]
\centering
\includegraphics[scale=0.6]{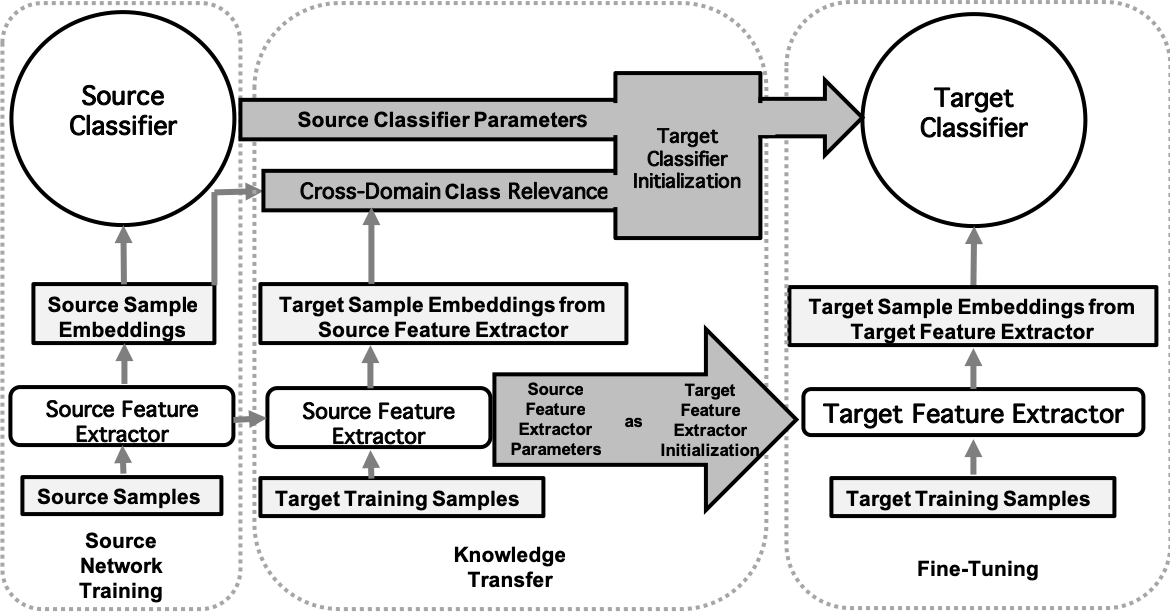}
\caption{\small \sl Graphical summary of the FSHAR framework.}
\label{harframework}
\end{figure}

\begin{figure}[t]
\centering
\includegraphics[scale=0.7]{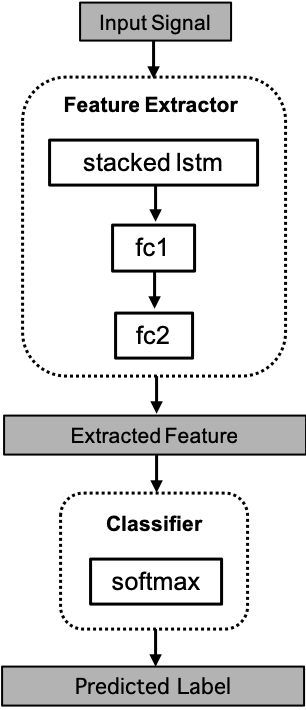}
\caption{\small \sl Network Structure}
\label{networkstructure}
\end{figure}

The basic framework for FSHAR is illustrated in Fig. \ref{harframework}. We first train a source network, which consists of a feature extractor and a classifier,  with source domain samples as training data and parameters being randomly initialized. The parameters of the source feature extractor and classifier are then separately transferred to a target network. For the source feature extractor, the parameters are copied to the corresponding part in the target network and are used as the initialization for the target feature extractor parameter optimization, a procedure known as fine-tuning\footnote{Fine-tuning is performed on both feature extractor and classifier of the target network.} in the literature. The transfer of source classifier parameters is dependent on a cross-domain class-wise relevance measure. The transferred classifier parameters are also used as the initialization for the target classifier. In this paper we use the same network structure for both source and target domains. The network structure is illustrated in Fig. \ref{networkstructure}. 

\subsection{Source Network Training}

We first train a source network with source domain samples to get a source feature extractor $f_{\rm src}(\cdot,\mathbf{\Theta}_{\rm src})$ with parameters $\mathbf{\Theta}_{\rm src},$\footnote{We often drop the dependence on $\mathbf{\Theta}_{\rm src}$ for readability, i.e., we use $f_{\rm src}(\cdot)$ to denote $f_{\rm src}(\cdot,\mathbf{\Theta}_{\rm src})$ with parameters $\mathbf{\Theta}_{\rm src}$ when no ambiguity is caused.} and a source classifier $C(\cdot,\mathbf{W}_{\rm src})$ with parameters $\mathbf{W}_{\rm src}$. The source classifier parameters can be represented as a matrix $\mathbf{W}_{\rm src} \in \mathbb{R}^{c_{\rm src} \times d}$, where $c_{\rm src}$ is the number of source classes and $d$ is the dimensionality of the encoded features $f_{\rm src}(\mathbf{X}_{\rm src}) \in \mathbb{R}^{n_{\rm src} \times d}$, which is used as the feature for classification, where $n_{src}$ denotes the number of source domain samples. We empirically use a stacked LSTM with two hidden layers followed by two fully connected layers as our feature extractor. By using LSTM, we can take advantage of the temporal dependencies within the HAR data, as the layout of an LSTM layer forms a directed cycle where the states of the network in current timestep depends on those of the network in the previous timestep. We use a softmax classifier due to its simplicity of training by gradient descent. \par

\subsection{Knowledge Transfer}
\label{trans}

We consider the information from the source feature extractor as ``generic information", as information from lower layers are believed to be more compatible between related tasks than that from higher layers\footnote{We define lower layers as layers that are more close to the input layer and higher layers as layers that are more close to the output layer of a network.} \cite{yosinski2014transferable}. To transfer the information from the source feature extractor, we copy the source feature extractor parameters $\mathbf{\Theta}_{\rm src}$ and use those as the initialization for the target feature extractor parameters, which means that the initialization of target feature extractor parameters $\mathbf{\Theta}_{\rm trg}^{0} = \mathbf{\Theta}_{\rm src}$. Since the information in a classifier is highly task-specific, it is necessary to pick information that is relevant to the target classes for knowledge transfer so that negative transfer can be alleviated. In our problem scenario, we focus on the class-wise relevance between the source and target training samples, which is referred as \emph{cross-domain class-wise relevance} in the sequel. We propose two different cross-domain class-wise relevance measures: one based on statistical relevance and another based on semantic relevance. \par
The statistical scheme to measure cross-domain class-wise relevance includes two steps. First, we calculate the cross-domain \emph{sample-wise} relevance, which measures the similarity between each pair of source domain sample and target training sample. Second, we calculate the cross-domain \emph{class-wise} relevance based on the obtained cross-domain \emph{sample-wise} relevance. Multiple distance metrics can be used to calculate cross-domain sample-wise relevance; we focus on two options for this work. The cross-domain sample-wise relevance values are stored in a matrix $\mathbf{A} \in \mathbb{R}^{n_{\rm src} \times n_{\rm trg}}$, where $n_{\rm trg}$ denotes the number of target training samples.
\begin{itemize}
\item{\textbf{Cosine Similarity}}, which uses the exponential value of the cosine similarity \cite{vinyals2016matching} to measure cross-domain sample-wise relevance. To be more specific, the relevance between the $i^\textrm{th}$ source domain sample and the $j^\textrm{th}$ target training sample is measured through  
\begin{equation}
\mathbf{A}^{(i,j)} = e^{[\tilde{f}_{\rm src}(\mathbf{X}_{\rm src}^{(i)})]^T\tilde{f}_{\rm src}(\mathbf{X}_{\rm trg}^{(j)})},
\end{equation} 
for $i=1,2,\cdots,n_{\rm src}$ and $j=1,2,\cdots,n_{\rm trg}$, where $\tilde{f}_{\rm src} (\cdot)=f_{\rm src} (\cdot)/\|f_{\rm src} (\cdot)\|_2$ denotes the normalized encoded feature. The exponential operation makes all relevance values positive to facilitate subsequent steps.
\item{\textbf{Sparse Reconstruction}}, which uses the magnitudes of the reconstruction coefficients to measure cross-domain sample-wise relevance under the assumption that there exists a linear mapping between source and target embeddings provided by the source feature extractor. That is, $f_{\rm src}(\mathbf{X}_{\rm trg}) = \mathbf{A}^Tf_{\rm src}(\mathbf{X}_{\rm src})$, where $\mathbf{A}$ acts as a reconstruction matrix with element values indicating cross-domain sample-wise relevance. We first solve the following minimization problem to get a reconstruction matrix $\mathbf{A}$:
\begin{equation}
\label{recon}
\min_{\mathbf{A}} \frac{1}{2n_{\rm trg}} \| \mathbf{A}^Tf_{\rm src}(\mathbf{X}_{\rm src}) - f_{\rm src}(\mathbf{X}_{\rm trg}) \|_F^2 + \lambda \| \mathbf{A} \|_{2,1},
\end{equation}
where $\lambda$ is a balance parameter. Since each row of $\mathbf{A}$ indicates the importance of the corresponding encoded source domain sample in reconstructing the encoded target training samples, we use the $\ell_2$-norm of each row of $\mathbf{A}$ to measure the relevance between an encoded source domain sample and the encoded target training samples, which leads to the $\ell_{2,1}$-norm regularization term in \eqref{recon} that enforces row sparsity on the transformation matrix $\mathbf{A}$ for similarity measure. 
\end{itemize}
With the obtained cross-domain sample-wise relevance matrix $\mathbf{A}$, we sum up the element values within each source-target class pair to get a class-wise relevance matrix $\mathbf{O} \in \mathbb{R}^{c_{\rm src} \times c_{\rm trg}}$, where $c_{\rm trg}$ is the number of target classes. That is, 
\begin{equation}
\mathbf{O}^{(p,q)} = \sum_{i \in s_p} \sum_{j \in s_q} \mathbf{A}^{(i,j)},
\end{equation}
for $p=1,2,\cdots,c_{\rm src}$ and $q=1,2,\cdots,c_{\rm trg}$, where $s_p$ and $s_q$ corresponds to the set of sample indices in the $p^{\textrm{th}}$ class and the $q^{\textrm{th}}$ class. We refer to the scheme with cosine similarity used for cross-domain sample-wise relevance measure as \textbf{FSHAR with Cosine Similarity (FSHAR-Cos)} and the one with sparse reconstruction used for cross-domain sample-wise relevance measure as \textbf{FSHAR with Sparse Reconstruction (FSHAR-SR)} in the sequel. \par
The semantic scheme to measure cross-domain class-wise relevance is based on the textual description of activity classes, in which multiple distance metrics can be used. In this paper, we employ the normalized Google distance (NGD) \cite{cilibrasi2007google} as the distance metric. NGD is a semantic similarity measure based on the number of hits returned by the Google search engine for a given pair of keywords. The NGD between keywords $P$ and $Q$ is calculated by
\begin{equation}
NGD(P,Q) = \frac{{\rm max} \{ \log g(P), \log g(Q) \} - \log g(P,Q)}{\log N -  {\rm min} \{ \log g(P), \log g(Q) \}},
\end{equation}
where $N$ is the total number of web pages searched by Google multiplied by the average number of search terms on each page, which is estimated by the number of hits by searching the word ``the"; $g(P)$ and $g(Q)$ are the number of hits for search terms $P$ and $Q$, respectively; and $g(P,Q)$ is the number of web pages on which both $P$ and $Q$ occur. Assume $P$ and $Q$ are the textual descriptions of the $p^\textrm{th}$ source class and the $q^\textrm{th}$ target class, respectively; then the cross-domain class-wise relevance matrix $\mathbf{O}$ is obtained through
\begin{equation}
\mathbf{O}^{(p,q)} = e^{-NGD(P,Q)}
\end{equation}
This scheme is referred to as \textbf{FSHAR with normalized Google distance (FSHAR-NGD)} in the sequel. \par
In order to facilitate classifier parameter transfer, we need to normalize the obtained cross-domain class-wise relevance matrix $\mathbf{O}$ such that for each target class the relevance values from all source classes sum to $1$. We propose two normalization schemes for comparison. The normalized cross-domain class-wise relevance values are stored in a matrix $\mathbf{W} \in \mathbb{R}^{c_{\rm src} \times c_{\rm trg}}$. 
\begin{itemize}
\item{\textbf{Scheme A}}: Soft normalization
\begin{equation}
\mathbf{W}^{(p,q)} = \frac{\mathbf{O}^{(p,q)}}{\sum_{p=1}^{c_{\rm src}} \mathbf{O}^{(p,q)}}
\end{equation}
\item{\textbf{Scheme B}}: Hard normalization
\begin{equation}
\mathbf{W}^{(p,q)} =
\begin{cases}
1,& \text{$\mathbf{O}^{(p,q)}=\{\max_{i}\mathbf{O}^{(i,q)}\}_{i=1}^{c_{\rm src}}$}\\
0,& \text{Otherwise}
\end{cases}
\end{equation}
\end{itemize}
The initial value of target classifier weights is a linear combination of the trained source classifier weights based on the normalized class-wise relevance matrix $\mathbf{W}$. That is, 
\begin{equation}
\label{wtrginit}
\mathbf{W}_{\rm trg}^0 = \mathbf{W}^T\mathbf{W}_{\rm src}, 
\end{equation}
where $\mathbf{W}_{\rm trg}^{0}$ denotes the initialization of target classifier parameters.  Compared with hard normalization, soft normalization may help improve knowledge transfer performance since it is able to capture the relationship between each single target class and multiple source classes instead of one. This is important for HAR tasks since there sometimes exists commonalities between activity categories. \par

\subsection{Fine-Tuning}

As mentioned earlier, both $\mathbf{\Theta}_{\rm trg}^0$ and $\mathbf{W}_{\rm trg}^0$ are used as the initialization for the feature extractor and the classifier parameters in the target network, respectively. The final target feature extractor parameter $\mathbf{\Theta}_{\rm trg}$ and classifier parameter $\mathbf{W}_{\rm trg}$ are obtained through backpropagation based fine-tuning with target training samples. The rationale behind fine-tuning is that not every target class has an unimodal distribution in the embedding space created by the source feature extractor as it could be biased towards features that are salient and discriminative among source classes, while fine-tuning is likely to move the embedding space so that each target class may have an unimodal distribution \cite{qi2018low}. FSHAR is summarized in Algorithm \ref{fshar}. \par
\begin{algorithm}[tbp]
\caption{\small \sl FSHAR}
\label{fshar}
\begin{algorithmic}[1]
\Require Source domain samples $\mathbf{X}_{\rm src}$; 
target training samples $\mathbf{X}_{\rm trg}$; 
\Ensure Target feature extractor parameters $\mathbf{\Theta}_{\rm trg}$; target classifier parameters $\mathbf{W}_{\rm trg}$.
\State Train a source network with source domain samples $\mathbf{X}_{\rm src}$ to obtain the source feature extractor parameters $\mathbf{\Theta}_{\rm src}$ and the source classifier parameters $\mathbf{W}_{\rm src}$;
\State Initialize the target feature extractor parameters $\mathbf{\Theta}_{\rm trg}^0$ using the source feature extractor parameters $\mathbf{\Theta}_{\rm src}$;
\State Calculate the normalized cross-domain class-wise relevance matrix $\mathbf{W}$ by following one of the schemes proposed in Section \ref{trans};
\State Initialize the target classifier parameters $\mathbf{W}_{\rm trg}^0$ by following Eq. \eqref{wtrginit};
\State Fine-tune the initialized target network to get the target feature extractor parameters $\mathbf{\Theta}_{\rm trg}$ and the target classifier parameters $\mathbf{W}_{\rm trg}$.
\end{algorithmic}  
\end{algorithm} 

\subsection{Implementation}

We used PyTorch \cite{paszke2017automatic} to implement the described framework. Network parameter optimization was performed via Adam \cite{kingma2014adam}. We employ the method proposed in \cite{liu2009multi}\footnote{Codes available at: \url{https://github.com/jundongl/scikit-feature/blob/master/skfeature/function/sparse_learning_based/ls_l21.py}} to solve the optimization problem \eqref{recon} , which efficiently solves an $\ell_{2,1}$-norm optimization problem by reformulating it as two equivalent smooth convex optimization problems.

\subsection{Limitation}
\label{limit}

As described in Section \ref{trans}, FSHAR uses a linear combination of source classifier parameters to get an initialization of classifier weights for target classes based on a normalized cross-domain class-wise relevance matrix $\mathbf{W}$. Although the matrix $\mathbf{W}$ only characterizes the relative relevance of source classes to each target class within the given source domain, it is also possible that none of the source classes is sufficiently relevant to a certain target class. In that case, FSHAR does not consider the domain-wise relevance for knowledge transfer. This limitation of FSHAR will be considered in our future work.

\subsection{Extension}
\label{ext}

As mentioned in Section \ref{intro}, there would be a practical impact if the knowledge extracted from new environments can be merged into the previously trained HAR system. In FSHAR, it is possible to merge the target model with the source model by concatenating the initialized target classifier weights with the source classifier weights as the initialization of a combined classifier $\mathbf{W}_{\rm comb}^0 = [ \mathbf{W}_{\rm src}; \mathbf{W}_{\rm trg}^0 ]$ and use $\mathbf{\Theta}_{\rm src}$ as the initialization for the feature extractor $\mathbf{\Theta}_{\rm comb}^0$ in the combined network. Fine-tuning can be performed on the combined network with training data consisting of the source domain samples and target training samples to get the optimal values of $\mathbf{\Theta}_{\rm comb}$ and $\mathbf{W}_{\rm comb}$.

\section{Experiments}
\label{exp}

In this section, we evaluate the knowledge transfer performance of FSHAR. Experiments are conducted on two benchmark human activity datasets. We also compare FSHAR with other relevant state-of-the-art techniques. The classification rates on target testing samples are used as the metric to evaluate knowledge transfer performance. We do not include experimental results of merged network as introduced in Section \ref{ext} as we only focus on the knowledge transfer performance that is reflected by classification accuracy on target testing data instead of a merged system.

\subsection{Dataset Information}

We first provide the overall information of each dataset and introduce the source/target domain setup. We perform experiments on two benchmark datasets: the Opportunity activity recognition dataset (OPP) \cite{chavarriaga2013opportunity} and the PAMAP2 physical activity monitoring dataset (PAMAP2) \cite{reiss2012introducing}. OPP consists of common kitchen activities from 4 participants with wearable sensors. Data from 5 different runs are recorded for each participant with activities being annotated with 18 mid-level gesture annotations. Following \cite{hammerla2016deep}, we only keep data from sensors without any packet-loss, which includes accelerometer data from the upper limbs and the back, and complete IMU data from both feet. The resulting dataset has 77 dimensions. PAMAP2 consists of 12 household and exercise activities from 9 participants with wearable sensors. The dataset has 53 dimensions. For frame-by-frame analysis, a sliding window with a one-second duration and 50$\%$ overlap is performed and the resulting data are used as inputs to the system for both datasets.  In order to eliminate the side effects caused by imbalanced classes, we set the number of samples from each class to be the same within each dataset through random selection. We keep 202 samples and 129 samples for each class of OPP and PAMAP2 when used as source data, respectively. 

\subsection{Source/Target Split}

In PAMAP2, the small number of samples for each participant may negatively affect the knowledge transfer performance when used as the source data. Therefore, the 9 participants are partitioned into 3 groups with the purpose of alleviating potential negative influences caused by the small number of samples in each class. To be more specific, the first three participants are in the first group, the second three participants are in the second group, and the remaining three participants are in the third group. \par
For each dataset, we select a portion of the classes as the source classes and the remaining as the target classes. Details on the source/target class split are listed in Table \ref{src/tgt-act}.\footnote{For OPP, Doors 1-2 denote two different doors and Drawers 1-3 denote three different drawers. When using NGD to calculate cross-domain class-wise relevance, we considered the same activity mode affecting on different objects as the same class. For example, we used ``Open Door" for both ``Open Door 1" and ``Open Door 2" when computing NGD.} We test the knowledge transfer performance in two scenarios: $i$) the source data come from the same participant/group of participants as those of the target data; $ii$) the source data come from different participants/groups of participants from those of the target data. In the first scenario, the target domain includes target activity classes of one participant for OPP or one group of participants for PAMAP2, and the source domain includes source activity classes of the same participant for OPP or the same group of participants for PAMAP2. In the second scenario, the target domain includes target activity classes of one participant for OPP or one group of participants for PAMAP2, and the source domain includes source activity classes of the remaining participants for OPP or the remaining groups of participants for PAMAP2.  \par

\begin{table}
\small
\centering
\begin{tabular}{|c|l|l|}  
\hline
\diagbox{Dataset}{Split}  & Source Activities & Target Activities \\
\hline
        & Open Door 2 & Open Door 1 \\
        & Close Door 2 & Close Door 1 \\
        & Close Fridge & Open Fridge \\
        & Close Dishwasher & Open Dishwasher \\
OPP & Close Drawer 1 & Open Drawer 1 \\
        & Close Drawer 2 & Open Drawer 2 \\
        & Close Drawer 3 & Open Drawer 3 \\
        & Clean Table & \\
        & Drink from Cup & \\
        & Toggle Switch & \\
\hline
 	       & Lying & Sitting \\
                & Standing & Cycling \\
                & Walking & Nordic Walking \\
PAMAP2 & Running & Descending Stairs \\
                & Ascending Stairs & Ironing \\
                & Vacuum Cleaning & \\
                & Rope Jumping & \\
\hline
\end{tabular}
\caption{\small \sl Source/target split for activities}
\label{src/tgt-act}
\end{table}

\subsection{Baselines}

\begin{table}
\small
\centering
\begin{tabular}{|l|c|c|}  
\hline
Parameters & OPP & PAMAP2  \\
\hline
LSTM Layer & 2 & 2 \\
\hline
LSTM Hidden Size & 64 & 50 \\
\hline
FC1 Size & 64 & 50 \\
\hline
FC2 Size & 64 & 25 \\
\hline
\end{tabular}
\caption{\small \sl Network Structure for Both Datasets}
\label{netstruct}
\end{table}

We compare our proposed FSHAR methods with the following four baselines.\footnote{We do not compare FSHAR with meta-learning-based FSL methods such as \cite{vinyals2016matching, snell2017prototypical} due to the generally limited number of classes in human activity datasets.} Note that all baselines are performed with the network structure described in Section \ref{framework}. The parameters for neural networks are listed in Table \ref{netstruct}. 
\begin{itemize}
\item{\textbf{Random Initialization (RandInit)}}, which trains the designed network with target training data from scratch and network parameters are randomly initialized. No knowledge is transferred from the source network.
\item{\textbf{Feature Extractor Transfer + Softmax Classifier (FeTr+Softmax)}}, which only transfers the source feature extractor parameters as the initialization for target feature extractor. The softmax is used as classifier and fine-tuning is performed on the whole network.
\item{\textbf{Feature Extractor Transfer + Nearest Neighbor Classifier (FeTr+NN)}}, which uses a copy of the source feature extractor parameters as that for the target feature extractor. Then a nearest neighbor classifier is applied on the embeddings extracted from both target training and testing samples. Following \cite{vinyals2016matching}, we first calculated the similarity between a given encoded target testing sample $\mathbf{x}_{\rm TrgTe}$ and different encoded target training samples $\mathbf{X}_{{\rm trg}(i)}$ for $i = 1,2,\cdots,n_{\rm trg}$ through 
\begin{equation}
S(\mathbf{x}_{\rm TrgTe},\mathbf{X}_{{\rm trg}(i)}) = \frac{e^{[\tilde{f}_{\rm src}(\mathbf{x}_{\rm TrgTe})]^T\tilde{f}_{\rm src}(\mathbf{X}_{{\rm trg}(i)})}}{\sum_{k=1}^{n_{\rm trg}} e^{[\tilde{f}_{\rm src}(\mathbf{x}_{\rm TrgTe})]^T\tilde{f}_{\rm src}(\mathbf{X}_{{\rm trg}(k)})}},
\end{equation}
where $\tilde{\mathbf{v}} = \mathbf{v}/\|\mathbf{v}\|$ is a normalization of vector $\mathbf{v}$.
\item{\textbf{Imprinting}}: Qi et al. \cite{qi2018low} proposed a weight imprinting approach that adds classifier weights in the final softmax layer for unseen categories by using a copy of the mean of the embedding layer activations extracted from the correponding training samples. Following \cite{qi2018low},we computed the classifier weights for a target class $c_j$ by averaging the embeddings $\mathbf{W}_{{\rm trg}(j)} = \frac{1}{t_j} \sum_{i=1}^{t_j} f_{\rm src} (\mathbf{X}_{{\rm trg}(i)})$, where $t_j$ is the number of samples in the $j^\textrm{th}$ target class. The obtained classifier weights were used as the initialization for the target classifier. Fine-tuning was then applied on the weights for optimization purposes.
\end{itemize}

\subsection{Performance Comparison}

\begin{table}
\small
\centering
\begin{tabular}{|c|c|c|c|c|c|c|c|c|c|}
\hline
\multicolumn{2}{|c|}{\multirow{2}{*}{\backslashbox{Model}{Participant}}} & \multicolumn{4}{|c|}{1-shot} & \multicolumn{4}{|c|}{5-shot} \\ \cline{3-10}
\multicolumn{2}{|c|}{} & 1 & 2 & 3 & 4 & 1 & 2 & 3 & 4 \\
\hline
\multicolumn{2}{|c|}{RandInit} & 37.06 & 39.39 & 32.21 & 40.48 & 54.56 & 56.99 & 53.87 & 60.46 \\
\hline
\multicolumn{2}{|c|}{FeTr+Softmax} & 47.52 & 46.83 & 42.37 & 41.08 & 63.00 & 62.18 & 59.19 & 57.83 \\
\multicolumn{2}{|c|}{FeTr+NN \cite{vinyals2016matching}} & 48.18 & 42.88 & 54.48 & 50.43 & 57.15 & 49.64 & 62.97 & 58.87 \\
\multicolumn{2}{|c|}{Imprinting \cite{qi2018low}} & 48.12 & 48.56 & 52.89 & 49.43 & 66.02 & 63.00 & \textbf{67.73} & 65.30 \\
\hline
\multirow{2}{*}{FSHAR-NGD} & Soft & \textbf{55.92} & \textbf{53.24} & \textbf{58.62} & \textbf{54.75} & \textbf{67.08} & \textbf{64.50} & \textbf{67.55} & \textbf{68.21} \\ 
& Hard & 51.96 & 50.48 & 55.44 & \textbf{54.20} & 61.99 & 60.73 & \textbf{66.95} & 67.70 \\
\multirow{2}{*}{FSHAR-Cos} & Soft & 52.53 & 49.87 & 55.82 & 52.54 & \textbf{66.66} & 62.74 & \textbf{67.68} & \textbf{69.05} \\ 
& Hard & 50.81 & 47.95 & 54.75 & 51.19 & 63.49 & 61.12 & \textbf{67.44} & 66.98 \\
\multirow{2}{*}{FSHAR-SR} & Soft & 50.53 & 47.71 & 54.49 & 52.42 & \textbf{66.19} & 61.61 & \textbf{67.64} & 67.89 \\
& Hard & 50.01 & 47.21 & 54.47 & 50.71 & 62.79 & 59.79 & \textbf{67.17} & 66.01 \\
\hline
\end{tabular}
\caption{\small \sl \small \sl Performance of FASTL and competing algorithms in classification on OPP where both source and target data come from the same participant. Classification accuracy (\%) on target testing samples is used as the evaluation metric.}
\label{opphom}
\end{table}

\begin{table}
\small
\centering
\begin{tabular}{|c|c|c|c|c|c|c|c|c|c|}
\hline
\multicolumn{2}{|c|}{\multirow{2}{*}{\backslashbox{Model}{Participant}}} & \multicolumn{4}{|c|}{1-shot} & \multicolumn{4}{|c|}{5-shot} \\ 
\cline{3-10}
\multicolumn{2}{|c|}{} & 1 & 2 & 3 & 4 & 1 & 2 & 3 & 4 \\
\hline
\multicolumn{2}{|c|}{RandInit} & 37.06 & 39.39 & 32.21 & 40.48 & 54.56 & 56.99 & 53.87 & \textbf{60.46} \\
\hline
\multicolumn{2}{|c|}{FeTr+Softmax} & 43.35 & 39.96 & 37.68 & 43.11 & 57.73 & 55.29 & 54.36 & 57.90 \\
\multicolumn{2}{|c|}{FeTr+NN \cite{vinyals2016matching}} & 45.91 & 37.43 & 32.99 & 41.09 & 53.61 & 46.11 & 37.77 & 42.58 \\
\multicolumn{2}{|c|}{Imprinting \cite{qi2018low}} & 45.94 & 42.25 & 42.04 & 44.17 & \textbf{61.45} & 59.33 & 60.85 & 59.17 \\
\hline
\multirow{2}{*}{FSHAR-NGD} & Soft & \textbf{47.79} & 48.07 & 47.63 & \textbf{49.06} & 61.17 & \textbf{61.88} & \textbf{62.08} & \textbf{60.64} \\
& Hard & \textbf{47.62} & \textbf{49.70} & \textbf{48.82} & \textbf{48.50} & 58.34 & 59.55 & 57.51 & 59.25 \\
\multirow{2}{*}{FSHAR-Cos} & Soft & \textbf{48.02} & 43.94 & 41.48 & \textbf{48.06} & \textbf{62.29} & \textbf{60.08} & 58.92 & 58.94 \\
& Hard & \textbf{47.17} & 42.54 & 39.54 & 46.72 & 59.74 & 58.80 & 55.99 & 57.43 \\
\multirow{2}{*}{FSHAR-SR} & Soft & \textbf{47.65} & 43.02 & 39.90 & 46.88 & \textbf{62.19} & 59.53 & 58.96 & 58.57 \\
& Hard & 45.65 & 40.56 & 37.09 & 46.37 & 59.38 & 55.58 & 54.72 & 58.61 \\
\hline
\end{tabular}
\caption{\small \sl Performance of FASTL and competing algorithms in classification on OPP where source and target data come from different participants. Classification accuracy (\%) on target testing samples is used as the evaluation metric.}
\label{opphet}
\end{table}

\begin{table}
\centering
\small
\begin{tabular}{|c|c|c|c|c|c|c|c|}
\hline
\multicolumn{2}{|c|}{\multirow{2}{*}{\backslashbox{Model}{Group}}} & \multicolumn{3}{|c|}{1-shot} & \multicolumn{3}{|c|}{5-shot} \\
\cline{3-8}
\multicolumn{2}{|c|}{\multirow{2}{*}{}} & 1 & 2 & 3 & 1 & 2 & 3 \\
\hline
\multicolumn{2}{|c|}{RandInit} & 41.20 & 38.36 & 49.73 & 50.62 & 53.00 & 64.65 \\
\hline
\multicolumn{2}{|c|}{FeTr+Softmax} & 42.97 & 49.34 & 42.90 & 54.08 & 55.43 & 54.72 \\
\multicolumn{2}{|c|}{FeTr+NN \cite{vinyals2016matching}} & 44.78 & 51.08 & 43.17 & 49.18 & 56.15 & 47.53 \\
\multicolumn{2}{|c|}{Imprinting \cite{qi2018low}} & 46.83 & 54.35 & \textbf{58.50} & 60.25 & 60.48 & \textbf{70.08} \\
\hline
\multirow{2}{*}{FSHAR-NGD} & Soft & 46.33 & 54.84 & \textbf{58.98} & 61.17 & 57.21 & 67.35 \\
& Hard & 44.64 & \textbf{57.67} & 47.90 & 51.93 & \textbf{62.84} & 61.12 \\
\multirow{2}{*}{FSHAR-Cos} & Soft & 48.29 & 56.00 & 56.83 & 62.74 & 60.52 & 68.97 \\
& Hard & 48.39 & 55.46 & 55.52 & \textbf{65.70} & 61.65 & 67.07 \\
\multirow{2}{*}{FSHAR-SR} & Soft & \textbf{50.87} & 55.07 & 55.25 & \textbf{65.43} & \textbf{62.45} & 68.75 \\
& Hard & 48.98 & 54.85 & 56.62 & \textbf{65.95} & \textbf{62.01} & 66.75 \\
\hline 
\end{tabular}
\caption{\small \sl Performance of FASTL and competing algorithms in classification on PAMAP2 where both source and target data come from the same group of participants. Classification accuracy (\%) on target testing samples is used as the evaluation metric.}
\label{pahom}
\end{table}

\begin{table}
\centering
\small
\begin{tabular}{|c|c|c|c|c|c|c|c|}
\hline
\multicolumn{2}{|c|}{\multirow{2}{*}{\backslashbox{Model}{Group}}} & \multicolumn{3}{|c|}{1-shot} & \multicolumn{3}{|c|}{5-shot} \\
\cline{3-8}
\multicolumn{2}{|c|}{\multirow{2}{*}{}} & 1 & 2 & 3 & 1 & 2 & 3 \\
\hline
\multicolumn{2}{|c|}{RandInit} & 41.20 & 38.36 & 49.73 & 50.62 & 53.00 & 64.65 \\
\hline
\multicolumn{2}{|c|}{FeTr+Softmax} & 41.33 & 43.91 & 56.27 & 52.24 & \textbf{59.22} & 71.80 \\
\multicolumn{2}{|c|}{FeTr+NN \cite{vinyals2016matching}} & 44.48 & 44.81 & 52.93 & 48.28 & 48.27 & 59.17 \\
\multicolumn{2}{|c|}{Imprinting \cite{qi2018low}} & 50.33 & 48.77 & \textbf{63.03} & 61.99 & 57.61 & 71.00 \\
\hline
\multirow{2}{*}{FSHAR-NGD} & Soft & 51.45 & \textbf{50.42} & \textbf{63.00} & 64.37 & 56.95 & 69.53 \\
& Hard & \textbf{54.59} & 48.93 & 62.37 & \textbf{67.24} & 53.51 & 65.48 \\
\multirow{2}{*}{FSHAR-Cos} & Soft & 50.72 & \textbf{50.58} & \textbf{62.82} & 63.83 & \textbf{58.58} & 74.28 \\
& Hard & 49.64 & \textbf{50.10} & 60.45 & 62.24 & 57.77 & 63.65 \\
\multirow{2}{*}{FSHAR-SR} & Soft & 50.77 & \textbf{50.44} & \textbf{63.70} & \textbf{67.91} & \textbf{58.56} & \textbf{77.18} \\
& Hard & 50.28 & 49.56 & 61.12 & 66.69 & 57.28 & 69.17 \\
\hline 
\end{tabular}
\caption{\small \sl Performance of FASTL and competing algorithms in classification on PAMAP2 wheresource and target data come from different groups of participants. Classification accuracy (\%) on target testing samples is used as the evaluation metric.}
\label{pahet}
\end{table}

We present the classification accuracy results of FSHAR and baselines on both datasets in Tables \ref{opphom}-\ref{pahet},\footnote{For FSHAR-SR, we conducted experiments with the balance parameter given ranges of $\lambda \in \{ 10^{-4}, 10^{-3}, 10^{-2}, 10^{-1}, 1 \}$. Since we observed that these values provided similar results, we only present results generated with balance parameter $\lambda=10^{-2}$ in Tables \ref{opphom}-\ref{pahet}, as we consider these results being representative of the performance of FSHAR-SR with a varying $\lambda$ value.} including different combinations of datasets (OPP or PAMAP2), whether source data and target data being generated from the same participant (OPP)/groups of participants (PAMAP2) or not, and the number of training samples from each target class (1 or 5). In these tables, each column corresponds to a participant (OPP)/group of participants (PAMAP2) under either the scenario of 1-shot or 5-shot. Each number is an average of results from 100 repetitions with varying target training samples. For each table, we highlight the best performance in each column as well as the classification accuracy values with a difference less than $1\%$ from the best one, since such a difference can be considered trivial. \par
It is obvious that the best performances are almost always achieved by FSHAR methods. More detailed analysis on different aspects is provided as below.
\begin{itemize}
\item{\textbf{Transfer Learning and Negative Transfer}}: We can find that most transfer learning based methods perform bettern than ``RandInit". However, we also see negative transfer when only the source feature extractor weights are transferred. We conjecture that the possible reason is that the weights we transferred are generated from all source domain samples without any regard for similarity measure and selection. Therefore, source domain samples with different relevance to the target domain make equal contributions to the weights and those with weak relevance may provide harmful information during training. The consistent poor performance of ``FeTr+NN" is possibly due to the unfeasibility for fine-tuning, which results in sub-optimal network parameter choices.
\item{\textbf{Number of Training Samples}}: Increasing the number of training samples for each target class from 1 to 5 can bring an increase of about 10$\%$ to 15$\%$ in performance for each method. It is also obvious that gap between ``RandInit", the baseline without knowledge transfer, and other transfer learning methods shrinks. We believe that by as the number of target training samples increase, the gap between performance of ``RandInit" and transfer learning based methods will keep shrinking, which decreases the necessity of transfer learning.
\item{\textbf{Source Data}}: For OPP, using source data from the same participant consistently provides better performance than that of using source data from different participants. Our method assumes that data from the same participants may have similar marginal distribution, though they have different conditional distribution due to disjoint label space between source domain and target domain. However, this is not true for PAMAP2. We conjecture that the combination of 3 participants into a group may introduce discrepancies in marginal distribution, which may leads to negative transfer.
\item{\textbf{Soft Normalization vs. Hard Normalization}}: Soft normalization does better than hard normalization in more cases than the other way around. For activity data, each activity may be correlated with multiple other activities instead of only a specific one. Soft normalization is able to characterize the relationship between a target activity with multiple source activities in the similarity, while hard normalization only selects the most similar source activity to each target activity, thus potentially degrading knowledge transfer performance due to the possible useful information from other source activities.
\item{\textbf{Cross-Domain Class-Wise Relevance Measures}}: For OPP, we find that FSHAR-Cos and FSHAR-SR provides similar performance in almost all cases. When each target class provides only one sample for training, FSHAR-NGD does better than both FSHAR-Cos and FSHAR-SR while when five samples are used, these three methods performs similarly. Due to the source/target split scheme we use especially for OPP, it is easy for NGD to find similarities between source and target classes using the semantic information in their labels, which also means the results from FSHAR-NGD may change when a different source/target split scheme is employed. This also explains the fact that the advantages of FSHAR-NGD is not significant for PAMAP2.  Furthermore, the performance of FSHAR-NGD may change if we use different texts to describe each activity class. Therefore, we argue that the advantages of FSHAR-NGD over FSHAR-Cos and FSHAR-SR can be a coincidence based on the source/target class split scheme and the way each activity class being described. But it is also noticeable that using statistical methods to calculate cross-domain class-wise relevance may suffer from insufficient training data, while the performance of semantic methods is independent of that.
\end{itemize}

\subsection{Effects of Fine-Tuning}

\begin{figure*}[!h]
\begin{minipage}{0.47\linewidth}
  \centerline{\includegraphics[width=6.5cm]{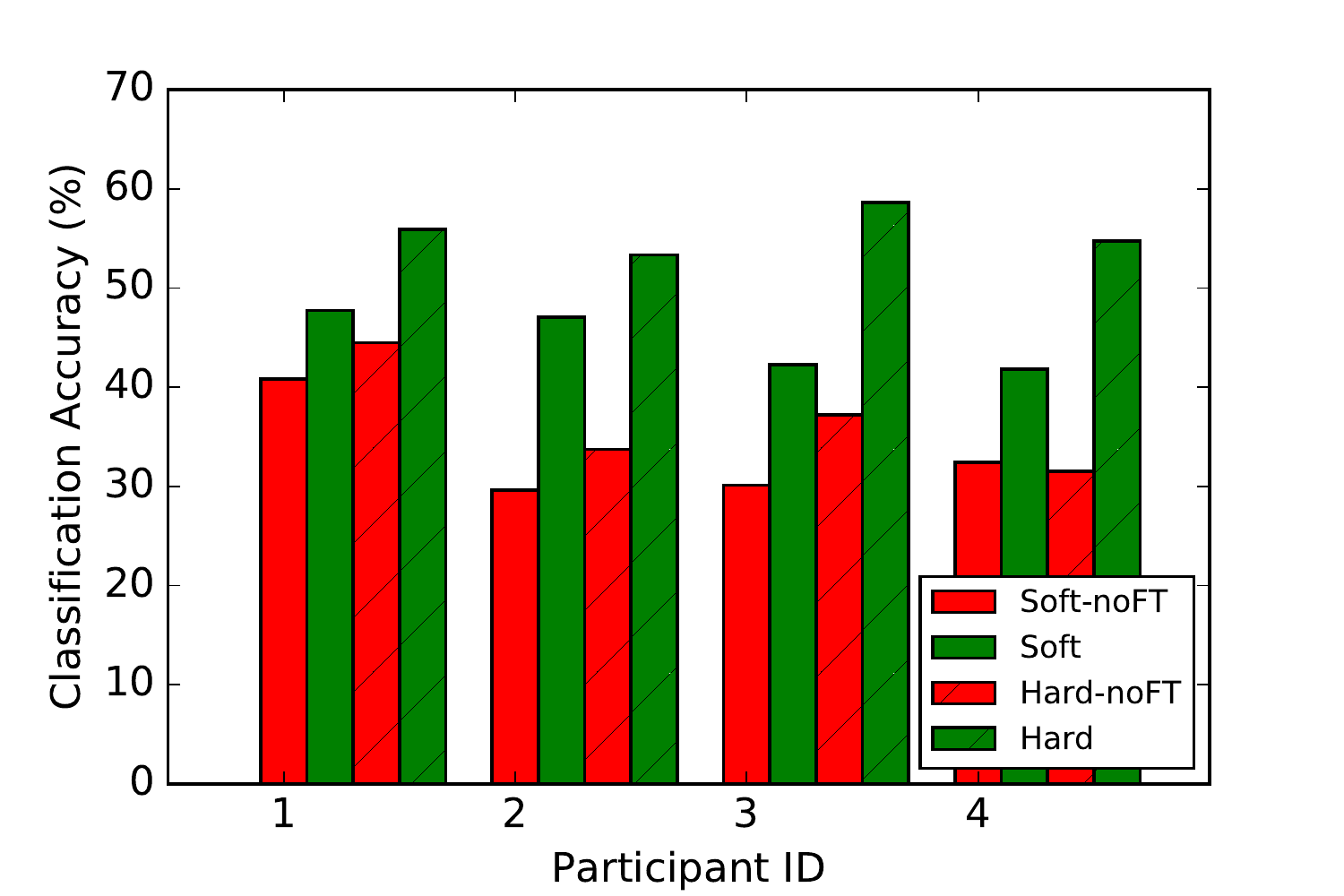}}
  \centerline{\small (a) 1-shot+FSHAR-NGD}
\end{minipage}
\hfill
\begin{minipage}{0.47\linewidth}
  \centerline{\includegraphics[width=6.5cm]{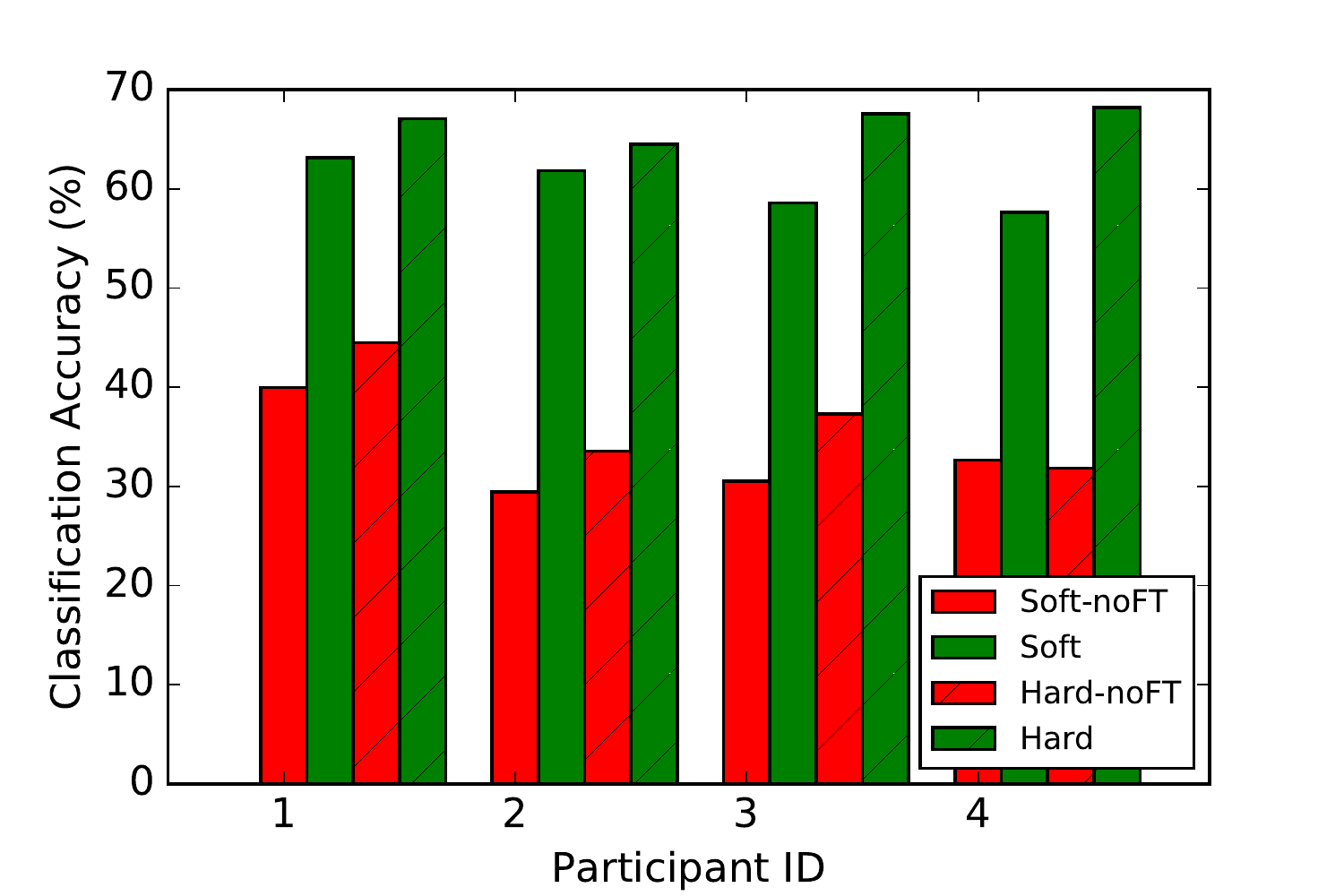}}
  \centerline{\small (b) 5-shot+FSHAR-NGD}
\end{minipage}
\vfill
\begin{minipage}{0.47\linewidth}
  \centerline{\includegraphics[width=6.5cm]{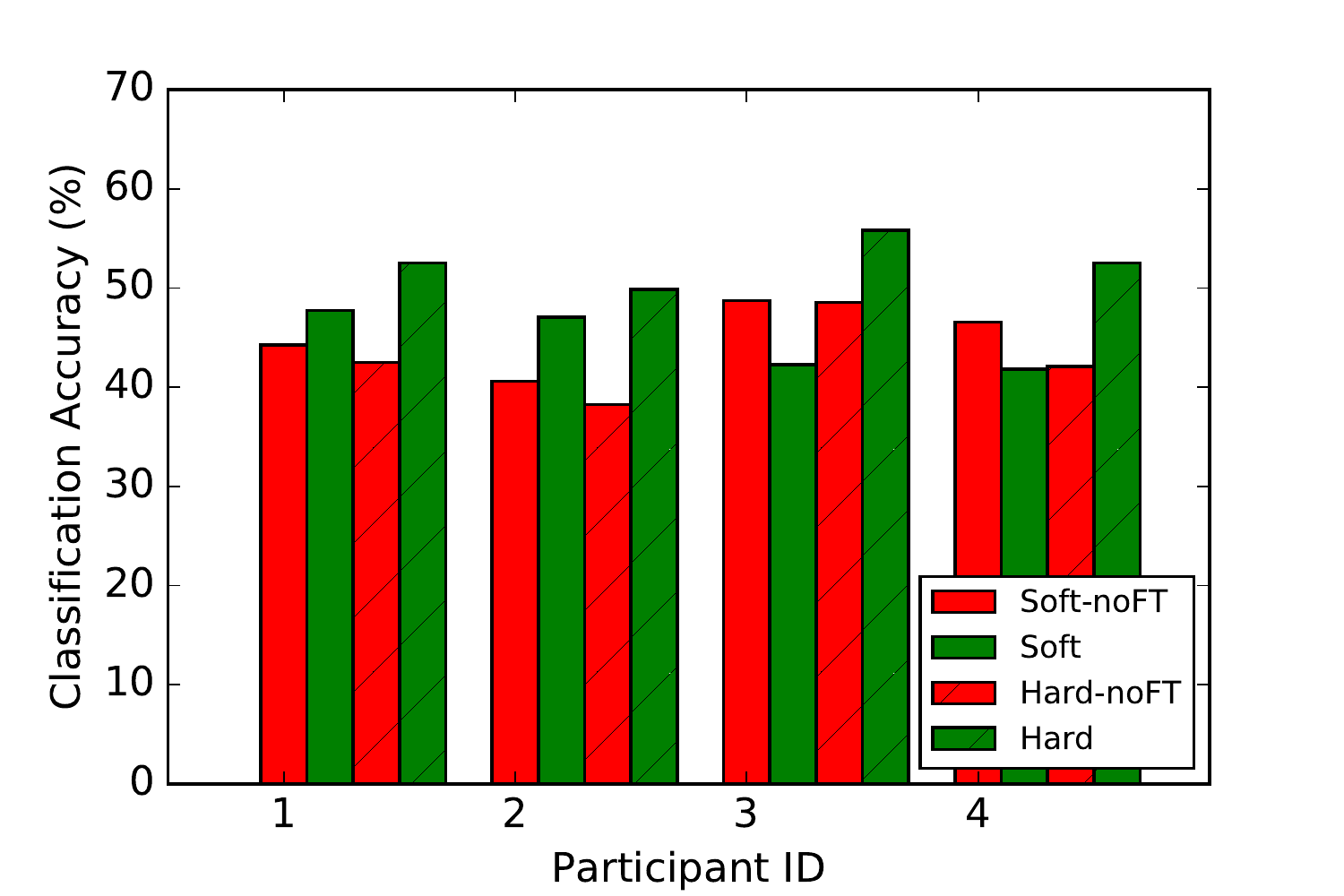}}
  \centerline{\small (c) 1-shot+FSHAR-Cos}
\end{minipage}
\hfill
\begin{minipage}{0.47\linewidth}
  \centerline{\includegraphics[width=6.5cm]{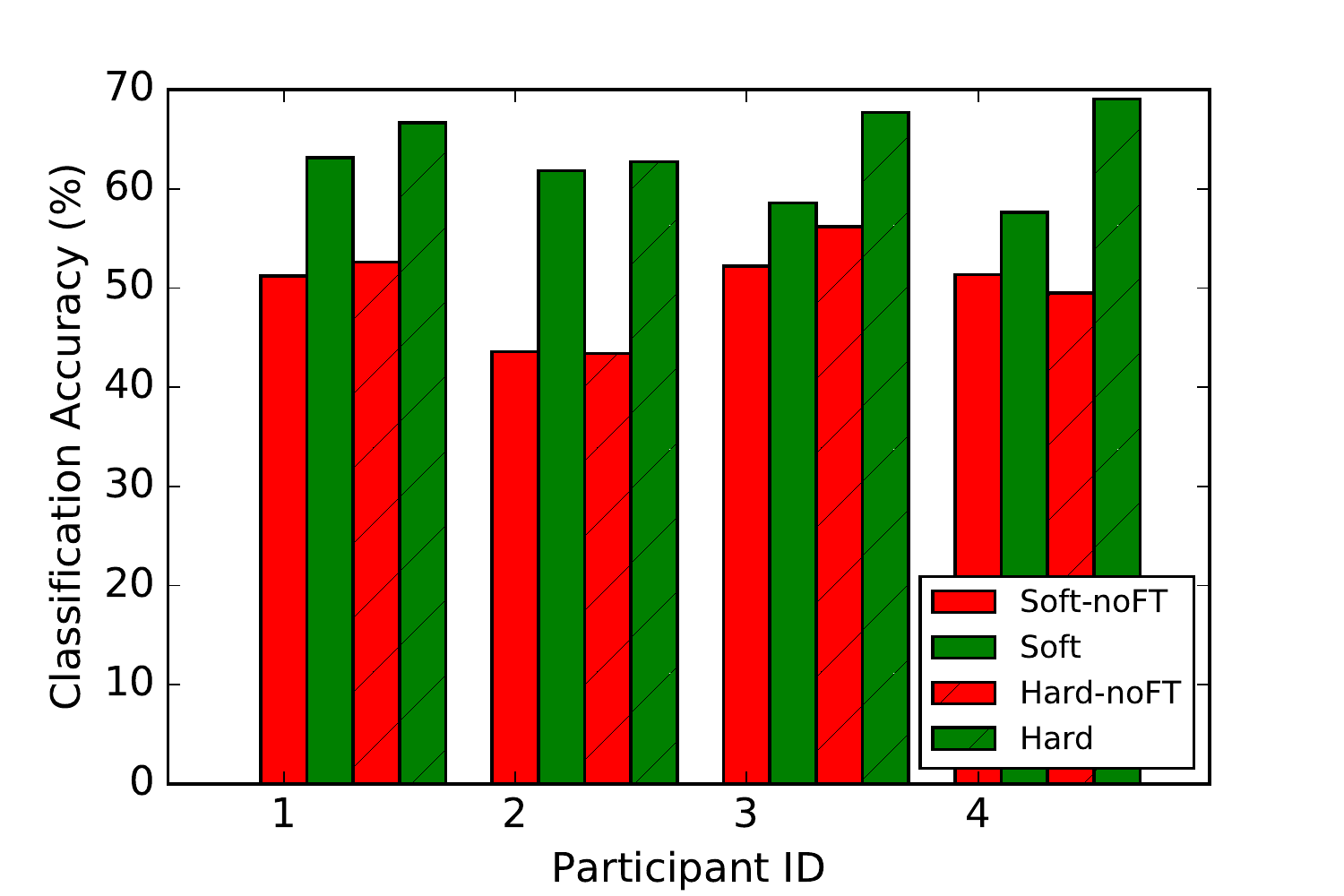}}
  \centerline{\small (d) 5-shot+FSHAR-Cos}
\end{minipage}
\vfill
\begin{minipage}{0.47\linewidth}
  \centerline{\includegraphics[width=6.5cm]{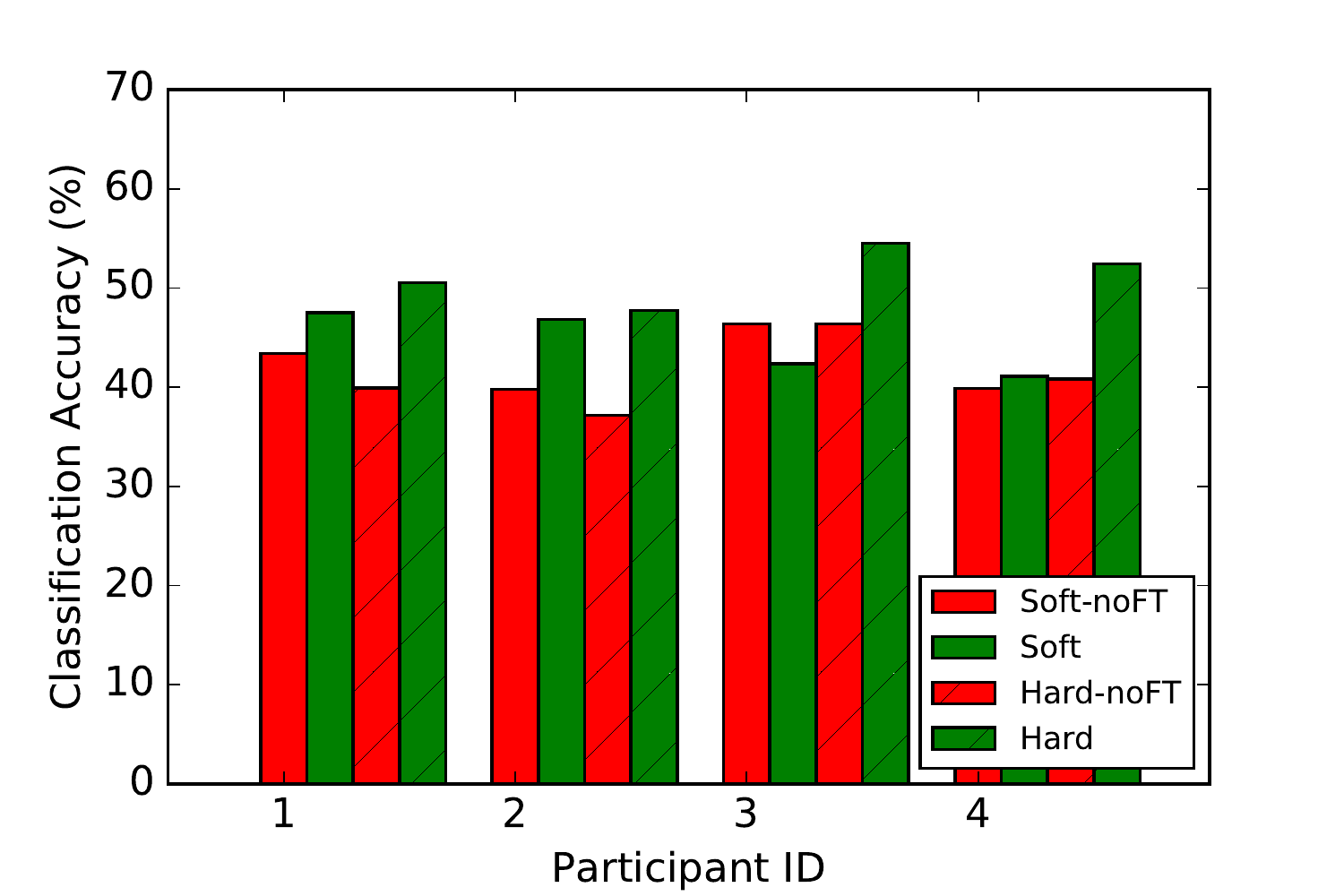}}
  \centerline{\small (e) 1-shot+FSHAR-SR}
\end{minipage}
\hfill
\begin{minipage}{0.47\linewidth}
  \centerline{\includegraphics[width=6.5cm]{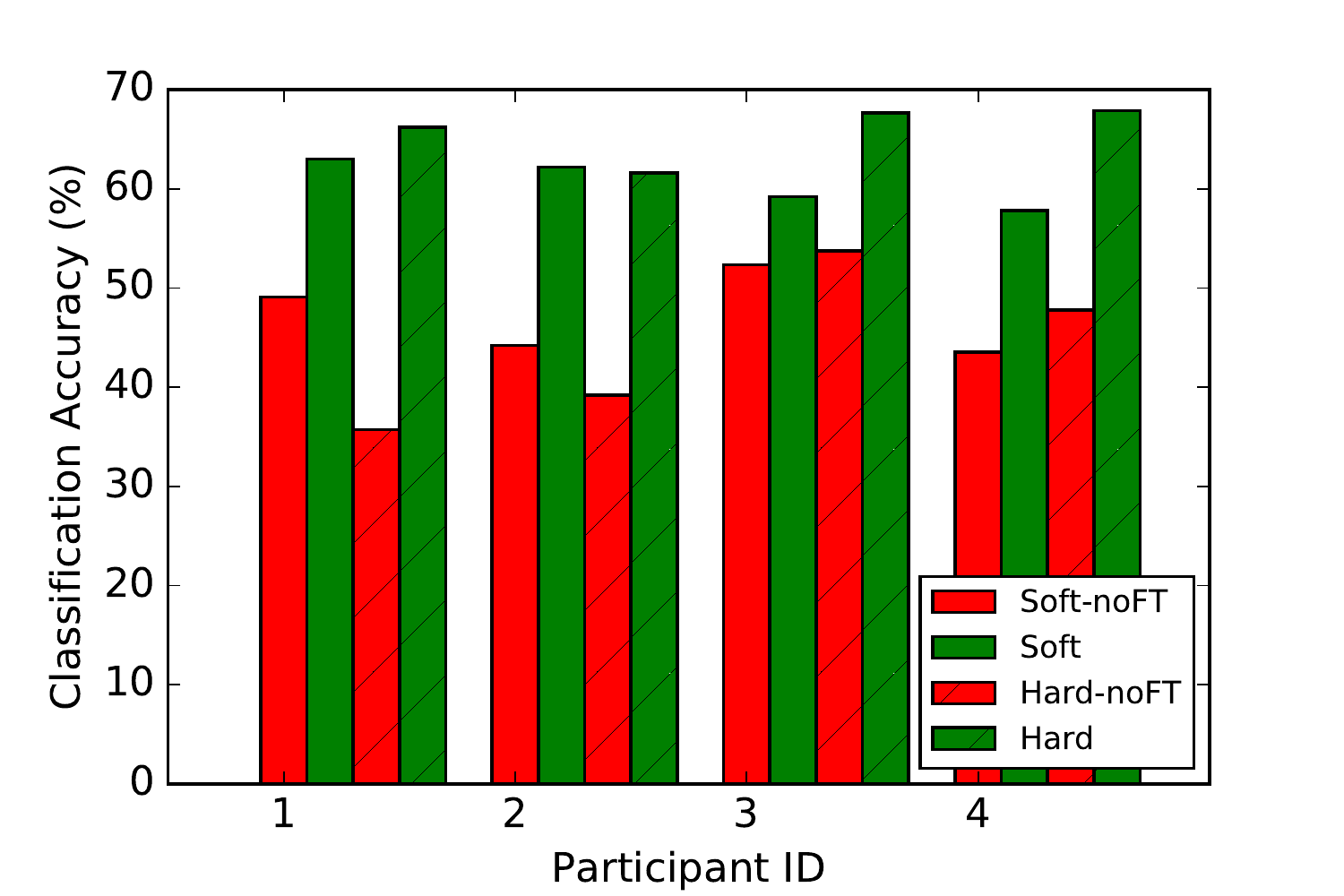}}
  \centerline{\small (f) 5-shot+FSHAR-SR}
\end{minipage}
\caption{\small \sl Effects of fine-tuning on OPP with source and target data coming from the same participant. Left column: 1-shot learning. Right column: 5-shot learning. Top row: FSHAR-NGD. Middle row: FSHAR-Cos. Bottom row: FSHAR-SR. In all subfigures, ``noFT" means without fine-tuning being performed on the initialized target network, and ``Soft"/``Hard" refer to soft/hard normalization scheme on the cross-domain class-wise relevance matrix $\mathbf{O}$ as described in Section \ref{trans}.}
\label{oppft}
\end{figure*}

We illustrate the effects of fine-tuning on the scenario of OPP with source and target data coming from the same participant in Fig. \ref{oppft}. It is obvious that fine-tuning improves classification performance in almost all cases except for the two of Participant 3 with the setup of 1-shot learning and soft normalization on the cross-domain class-wise relevance matrix for both FSHAR-Cos and FSHAR-SR. Additionally, the improvements become larger with the increase of number of training samples. We conjecture that with only one training sample from each target class, fine-tuning is more likely to lead to overfitting that degrade the generalization capability of the learned model. Results from other scenarios (i.e. PAMAP2, source and target data from different participant/group of participants) are not included due to the fact that they perform similar patterns and the purpose of saving space. 

\section{Conclusion}
\label{con}

We propose a few-shot learning framework for human activity recognition. The main contributions in this proposed approach is two-fold. First, we propose a method to transfer the parameters for a feature extractor and classifier from the source network to the target network based on cross-domain class-wise relevance. Second, the proposed framework is a general framework where different cross-domain class-wise relevance measure can be embedded. With the proposed framework, satisfying human activity recognition results can be achieved even when only very few training samples are available for each class. Experimental results show the advantages of the framework over methods with no knowledge transfer or that only transfer knowledge of feature extractor. \par
Future work includes four aspects: 1) we will study and try to overcome the problem of cross-domain relevance; 2) a merged model after knowledge transfer is also worth exploring; 3) we may combine statistical ways and semantic ways to measure cross-domain class-wise relevance instead of separating them; 4) it is worth trying to find a way to combine source data from both same and different participants from the target participant to see if knowledge transfer performance can be improved.

\end{document}